\documentclass[letterpaper, 10 pt, journal, twoside]{IEEEtran} 

\IEEEoverridecommandlockouts       

\PassOptionsToPackage{export}{adjustbox}
\usepackage{definitions}
\usepackage[T1]{fontenc}
\usepackage{tikz}
\usepackage{flushend}

\usepackage{dsfont}
\usepackage[cal=pxtx,scr=boondox]{mathalpha}
\usepackage{bm}
\usepackage{yhmath}
\usepackage{adjustbox}
\usepackage{mathtools}
\usepackage{amsmath}
\usepackage{amssymb} 
\usepackage[linesnumbered,ruled,vlined]{algorithm2e}
\usepackage{url}
\usepackage{graphics}
\usepackage{graphicx}
\usepackage{epsfig}
\usepackage{textcomp}
\usepackage{cite}
\usepackage{bm}
\usepackage{subcaption}
\usepackage[labelfont=bf]{caption}
\usepackage{booktabs}
\usepackage[english]{babel}
\usepackage[utf8]{inputenc}
\usepackage{theorem}
\usepackage{pifont}
\usepackage{cuted}
\usepackage{colortbl,xcolor}
\usepackage{multirow}
\usepackage{siunitx}

\usepackage{enumitem}

\newcommand{\spx}{\mathscr{x}}
\newcommand{\spr}{{\mathscr{r}}}
\newcommand{\sps}{{\mathscr{s}}}
\newcommand{\rg}[1]{{\mathring{#1}}}
\newcommand{\blam}{{\bm{\Lambda}}}
\newcommand{\bmo}{{\bm{\Omega}}}
\newcommand{\bmp}{{\bm{\Phi}}}

\newcommand{\qf}{{\mathscr{Q}}}

\newcommand{\fimu}{{\ttt{I}}}
\newcommand{\flid}{{\ttt{L}}}
\newcommand{\acce}{{\ttt{acc}}}
\newcommand{\gyro}{{\ttt{gyro}}}

\definecolor{kitred}{RGB}{187,25,23}

\begin{document}
	
	\markboth{PUBLISHED ON IEEE ROBOTICS AND AUTOMATION LETTERS (RA-L). DOI: 10.1109/LRA.2025.3604758. ©IEEE}
	{Cao \MakeLowercase{\textit{et al.}}: RESPLE: Recursive Spline Estimation for LiDAR-Based Odometry}
	
	\author{Ziyu~Cao$^{1}$, William Talbot$^{2}$, and Kailai~Li$^{3}$
	\thanks{Published on IEEE Robotics and Automation Letters (RA-L). ©IEEE. Personal use of this material is permitted. Permission from IEEE must be obtained for all other uses. DOI: 10.1109/LRA.2025.3604758. The work of William Talbot was supported by the Swiss National Science Foundation (SNSF) [20CH-1 229464/1] under CHIST-ERA grant CHIST-ERA-23-MultiGIS-07.}
	\thanks{$^{1}$Ziyu Cao is with the Department of Electrical Engineering, Linköping University, Sweden \tt{\footnotesize ziyu.cao@liu.se}}
	\thanks{$^{2}$William Talbot is with the Robotic Systems Lab (RSL), ETH Zurich, Switzerland \tt{\footnotesize wtalbot@ethz.ch}}
	\thanks{$^{3}$Kailai Li is with the Bernoulli Institute for Mathematics, Computer Science and Artificial Intelligence, University of Groningen, the Netherlands \tt{\footnotesize kailai.li@rug.nl}}
	\thanks{Digital Object Identifier (DOI): see top of this page.}
	}
	
	\title{RESPLE: Recursive Spline Estimation for LiDAR-Based Odometry}
	\maketitle
	
	\begin{abstract}
		We present a novel recursive Bayesian estimation framework using B-splines for continuous-time 6-DoF dynamic motion estimation. The state vector consists of a recurrent set of position control points and orientation control point increments, enabling efficient estimation via a modified iterated extended Kalman filter without involving error-state formulations. The resulting recursive spline estimator (RESPLE) is further leveraged to develop a versatile suite of direct LiDAR-based odometry solutions, supporting the integration of one or multiple LiDARs and an IMU. We conduct extensive real-world evaluations using public datasets and our own experiments, covering diverse sensor setups, platforms, and environments. Compared to existing systems, RESPLE achieves comparable or superior estimation accuracy and robustness, while attaining real-time efficiency. Our results and analysis demonstrate RESPLE's strength in handling highly dynamic motions and complex scenes within a lightweight and flexible design, showing strong potential as a universal framework for multi-sensor motion estimation. We release the source code and experimental datasets at \texttt{https://github.com/ASIG-X/RESPLE}.
	\end{abstract}
	
	
	\section{Introduction}\label{sec:intro}
	\IEEEPARstart{R}{eliable} estimation of dynamic egomotions using onboard sensors is critical for mobile robots to achieve high-performance autonomy in ubiquitous application scenarios, such as autonomous driving, service robotics, and search and rescue~\cite{tranzatto2022cerberus,miki2022learning}. Multi-sensor solutions involving LiDARs have gained significant popularity due to certain advantages in perception, including resilience to varying lighting conditions, spatiotemporally dense observations, high accuracy, and long detection range. Recent advances in lightweight, versatile designs, and improved cost-effectiveness have further fueled the adoption of LiDAR technology~\cite{RAL21_Li,xu2022fast}.
	\begin{figure}[t]
		\centering
		\adjustbox{trim={0.\width} {0.\height} {0.\width} {0.\height},clip}{\includegraphics[width=0.96\linewidth]{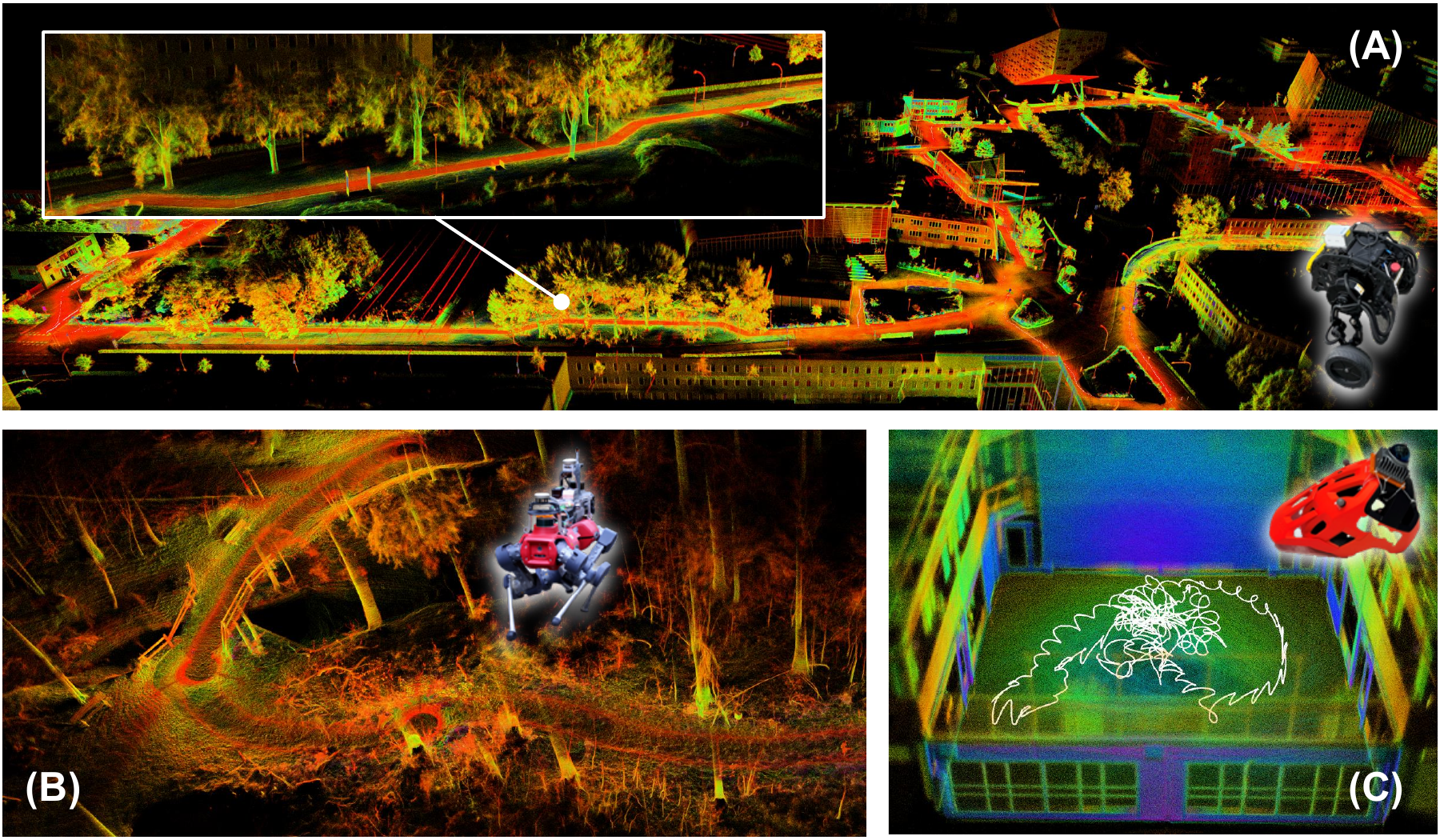}}
		\caption{RESPLE tested on (A) a wheeled bipedal robot in a campus environment, (B) a quadruped robot in the wild, and (C) a helmet platform indoor in highly dynamic motions.}
		\label{fig:front}
		\vspace{-5mm}
	\end{figure}
	
	Traditionally, dynamic motion estimation has been addressed in the discrete-time domain, where states are estimated often at a fixed rate via filtering or nonlinear optimization (smoothing). The former, such as the error-state Kalman filter, recursively predicts and updates state estimates according to predefined process and measurement models~\cite{sola2017quaternion,bloesch2017iterated}. The latter, often applied in a sliding-window fashion, optimizes states via maximum likelihood estimation (MLE) or maximum a posteriori (MAP) estimation~\cite{vins2018,RAL21_Li}. Both methodologies have well established solutions for LiDAR-based odometry that exploit efficient map representation, such as the ikd-Tree~\cite{cai2021ikd} and iVox~\cite{bai2022faster}, and per-point residuals, achieving real-time performance~\cite{xu2022fast,he2023point}. The inertial measurement unit (IMU) is commonly integrated for explicit motion compensation of LiDAR scans or better handling dynamic motions in general. However, in the presence of spatiotemporally dense and asynchronous observations from multiple sensors, discrete-time methods face limitations in high-fidelity processing, accommodating varying sensor rates, and maintaining estimation accuracy without incurring excessive theoretical or computational complexity~\cite{talbot2024continuous}.
	
	The continuous-time paradigm provides effective alternatives for estimating dynamic motion and is of increasing interest in LiDAR-based odometry for mobile robotics. Common choices of continuous-time motion models include piecewise-linear functions, Gaussian processes (GPs) and B-splines, whereby interpolated residuals can be computed for estimation~\cite{talbot2024continuous}. Piecewise-linear functions are well motivated by the constant-velocity assumption applied in LiDAR scan de-skewing~\cite{zhang2014loam} and have been embraced in recent LiDAR odometry systems~\cite{zheng2024trajlo}. However, this piecewise constant-velocity assumption may not  sufficiently capture dynamic motions in modern robotic systems, motivating an explorations of more expressive representations. Interpolation with `exactly sparse' GPs has emerged as an effective continuous-time estimation approach~\cite{barfoot2024state}. The interpolation relies on chosen motion models, typically constant-velocity/-acceleration, and has been applied in various LiDAR-based estimation pipelines~\cite{burnett2024continuous,shen2025cte}. B-splines (typically uniform and cubic) are popular in multi-sensor motion estimation and have demonstrated promising gains in accuracy and robustness in LiDAR-inertial odometry (LIO)~\cite{lv2021clins,ng2024eigen}. However, B-spline-based continuous-time LiDAR-only and multi-LiDAR odometry remain largely unaddressed in both methodological  and practical development.
	
	Most continuous-time estimation approaches adopt the strategy of sliding-window optimization incorporating multi-sensor interpolated residuals. This allows for direct incorporation of high-rate measurements at their exact timestamps and eliminates the need for motion compensation in LiDAR scans as preprocessing. However, such optimization-based designs often rely on highly performant, custom-built solvers, which pose significant challenges in computational efficiency and versatility, particularly in multi-sensor and mobile application scenarios~\cite{RAL23_Li,ng2024eigen}.
		
	In contrast, recursive Bayesian estimation offers a conceptually lightweight, computationally efficient, and pragmatic alternative that has been widely adopted in discrete-time LIO~\cite{xu2022fast,he2023point}, yet remains underexplored in the continuous-time paradigm. CTE-MLO~\cite{shen2025cte} presented a GP-based extended Kalman filter (EKF) for real-time multi-LiDAR odometry onboard common aerial and wheeled platforms. However, the adopted motion model assumes constant acceleration and angular velocity, which may limit expressiveness in capturing highly dynamic motions. \cite{li2025embedding} proposed a B-spline-embedded recursive estimation scheme in Euclidean spaces with limited validations on positioning using sensor networks. As such, it lacks applicability to full 6-DoF motion estimation involving orientations for LiDAR-based odometry. To the best of the authors' knowledge, no B-spline recursive estimator has been introduced to estimate 6-DoF dynamic motions, including for LiDAR-based odometry.
	
	\subsubsection*{Contributions} 	
	Motivated by the limitations of related work, we introduce RESPLE (\textbf{Re}cursive \textbf{Spl}ine \textbf{E}stimator) for universal LiDAR-based odometry. 
		\begin{itemize}[leftmargin=*]
			\item RESPLE is the first B-spline recursive estimation framework for estimating full 6-DoF dynamic motions. 6-DoF cubic B-splines are embedded into state-space modeling, where the state vector comprises a recurrent set of position control points and orientation control point increments. A modified iterated EKF is further proposed for efficient motion estimation without error-state formulations. 
			\item Using RESPLE as the estimation backbone, we develop a versatile suite of direct LiDAR, LiDAR-inertial, multi-LiDAR, and multi-LiDAR-inertial odometry systems within a unified system design.
			\item We conduct extensive real-world evaluations using public datasets and experiments across diverse application scenarios. Compared to existing systems, RESPLE achieves comparable or better performance in terms of estimation accuracy and robustness with real-time efficiency.
			\item RESPLE evidently demonstrates its strength in handling challenging conditions (e.g., highly dynamic motions and complex environments) with a lightweight design, highlighting its strong potential as a universal multi-sensor motion estimation framework. We publicly release our implementation and experimental datasets.
		\end{itemize}
	
	\section{Preliminaries}\label{sec:pre}
	\subsection{Notation Conventions}
	Throughout the following content, scalar values are written as lowercase letters, such as $a$. We use underlined lowercase letters, such as $\ua$, to denote vectors and bold capital letters, such as $\fA$, for  matrices. Continuous functions are denoted by italic letters, such as $\sps(t)$. Operators $\bullet$ and $\otimes$ are used to denote the Hamilton and Kronecker product, respectively.
	\subsection{Continuous-Time Parameterization of $6$-DoF Motions}\label{subsec:ctb}
	We exploit cubic B-splines (fourth-order) to represent $6$-DoF motions in the continuous-time domain~\cite{RAL23_Li} as follows
	\begin{equation}\label{eq:spl}
		\spx(t)=[\sps(t)^\top,\spr(t)^\top]^\top\in\R^3\times\Sbb^3\subset\R^7\,.
	\end{equation}
	$\sps(t)$ and $\spr(t)$ are the position and quaternion-valued orientation spline components, respectively, determined by the control points, $\{\us_i\}_{i=1}^n$ and $\{\ur_i\}_{i=1}^n$ over knots $\{t_i\}_{i=1}^n$ with a uniform temporal interval $\tau$. The separation of poses into their position and orientation components is supported in literature~\cite{ovren2019trajectory,hug2022con,RAL23_Li}, with the decoupling more computationally efficient and more appropriate for handheld and mobile robot motions. Given an arbitrary time instant $t\in[\,t_{n-1},t_{n})$, the position can be obtained according to
	\begin{equation}\label{eq:st}
		\sps(t)=[\,\us_{n-3},\us_{n-2},\us_{n-1},\us_{n}]\,\bom\,\uu\,, \eqwith
	\end{equation}
	\begin{equation}\label{eq:u}
		\bmo=\textstyle\frac{1}{6}\Bigg[\sbbmat\,1&-3&3&-1\,\\4&0&-6&3\\1&3&3&-3\\0&0&0&1\sebmat\Bigg]\eqand\uu=[\,1,u,u^2,u^3]^\top
	\end{equation}
	denoting the basis matrix and powers of normalized time $u=(t-t_{n-1})/\tau$, respectively. Similarly, the quaternion-valued B-spline in \eqref{eq:spl} at $t\in[\,t_{n-1},t_{n})$ takes the following cumulative expression
	\begin{equation}\label{eq:rt}
		\textstyle\spr(t)=\ur_{n-4}\bullet\prod_{i=n-3}^n\Exp_\dso\big(\lambda_i\ude_i\big)\,.
	\end{equation} 
	$\ude_i$ is the increment of adjacent control points measured in the tangent space with
	\begin{equation}\label{eq:ude}
		\ude_i=\Log_\dso(\ur_{i-1}^{-1}\bullet\ur_i)\,,\quad \text{for}\quad i = n-3,\cdots,{n}\,.
	\end{equation}
	$\Log_\dso(\cdot)$ and $\Exp_\dso(\cdot)$ are the logarithm and exponential maps at identity quaternion $\dso=[\,1,0,0,0\,]^\top$~\cite{sola2017quaternion}. In accordance with the cumulative B-spline in \eqref{eq:rt}, the cumulative basis functions $\{\lambda_i\}_{i=n-3}^n$ are given by
	\begin{equation}\label{eq:uLam}
		[\,\lambda_{n-3},\lambda_{n-2},\lambda_{n-1},\lambda_n\,]^\top=\bmp\,\uu\,,\eqwith
	\end{equation}
	$\bmp=\textstyle\frac{1}{6}\Bigg[\sbbmat\,6&0&0&0\\5&3&-3&1\,\\1&3&3&-2\\0&0&0&1\sebmat\Bigg]$ being the cumulative basis matrix~\cite{RAL23_Li}. 
	
	\subsection{Kinematic Interpolations}\label{subsec:kinoItp}
	We now present kinematic interpolations on the 6-DoF B-spline in IMU-involved multi-sensor settings, including temporal derivatives of the position and orientation components up to the second and the first order, respectively.
	
	\subsubsection{Positions}\label{subsec:transItp}
	The position B-spline in \eqref{eq:st} has a linear expression w.r.t.\ the normalized time vector $\uu$. According to~\cite{li2025embedding}, it is straightforward to derive the following generic expression for position kinematics via vectorization of \eqref{eq:st}
	\begin{equation}\label{eq:stKino}
		\begin{aligned}
			\rg{\sps}(t)&=\rg{\blam}\,[\,\us_{n-3}^\top,\us_{n-2}^\top,\us_{n-1}^\top,\us_{n}^\top]^\top\,,\eqwith\\
			\rg{\blam}&=(\bmo\,\rg{\uu})^\top\otimes\fI_3\in\R^{3\times12}\,.
		\end{aligned}
	\end{equation} 
	`$\circ$' serves as an umbrella symbol for the zeroth- to second-order temporal derivatives of the function underneath, such as position $\sps(t)$, velocity $\dot{\sps}(t)$, and acceleration $\ddot{\sps}(t)$. Derivatives of the normalized time vector ${\uu}$ can be derived given~\eqref{eq:u}.
	
	\subsubsection{Orientations}\label{subsec:rotItp} 
	In accordance with gyroscope observations, we provide the first-order temporal derivative of the orientation B-spline \eqref{eq:rt}, namely, the angular velocity function $\uom(t)$ w.r.t.\ the body frame. This follows the recursive computation procedure presented in~\cite{RAL23_Li}
	\begin{equation}\label{eq:omrec}
		\begin{aligned}
			\uom_1(t)&=2\dot{\lambda}_{n-2}\,\ude_{n-2},\\
			\uom_2(t)&=\ue_{n-1}^{-1}\bullet\uom_1(t)\bullet\ue_{n-1}+2\dot{\lambda}_{n-1}\,\ude_{n-1}\,,\\
			\uom(t)&=\ue_n^{-1}\bullet\uom_2(t)\bullet\ue_{n}+2\dot{\lambda}_{n}\,\ude_{n}\,,\\
		\end{aligned}
	\end{equation}
	with $\ue_i=\Exp_{\mathds{1}}{(\lambda_i\,\ude_i)}$, for $i=n-1$ and $n$. The derivatives of the cumulative basis functions in \eqref{eq:uLam} are given by $[\,\dot{\lambda}_{n-3},\dot{\lambda}_{n-2},\dot{\lambda}_{n-1},\dot{\lambda}_n\,]^\top=\bmp\,\dot{\uu}$.
	
	\section{Recursive motion estimation on B-Splines}\label{sec:resple}
	\begin{figure}[t]
		\vspace{1mm}
		\centering
		\adjustbox{trim={0.0\width} {0.\height} {0.0\width} {0.\height},clip}{\includegraphics[width=1\linewidth]{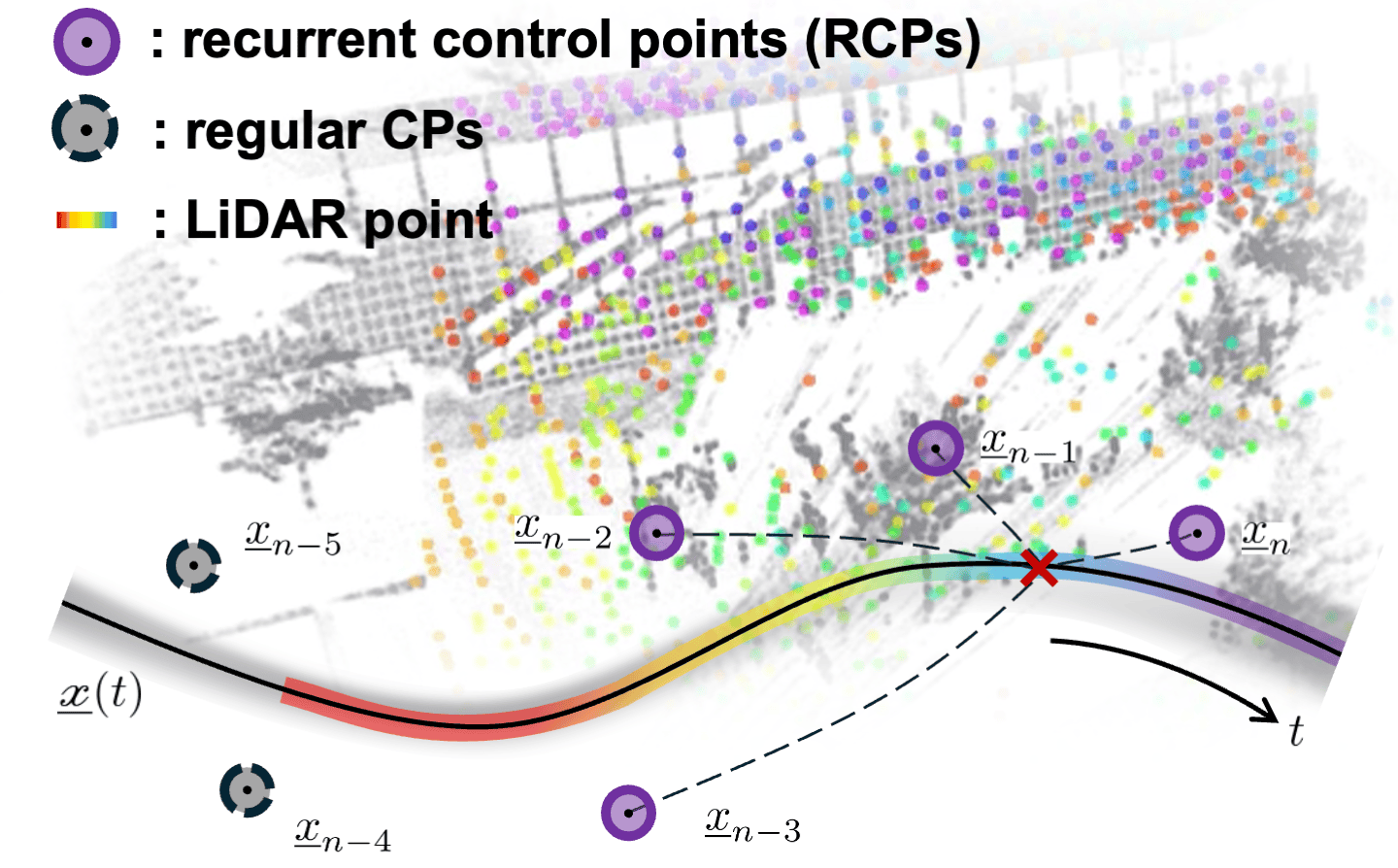}}
		\caption{Conceptual illustration of RESPLE-LO. LiDAR points with exact timestamps (colored) recursively update B-spline trajectory (black curve) with uncertainties (gray band).}
		\vspace{-4mm}
		\label{fig:spline}
	\end{figure}
	
	\subsection{6-DoF Spline-State-Space (TriS) Model}\label{subsec:6tris}
	We extend the basic spline-state-space modeling introduced in~\cite{li2025embedding} from Euclidean-only motion to the complete 6-DoF motion representation. Concretely, the state vector follows $\ux_k=[\,(\ux_k^\tts)^\top,(\ux_k^\ttr)^\top\,]^\top\in\R^{24}$, with 
	\begin{equation}\label{eq:rcp}
		\begin{aligned}
			\ux^\tts_k&=[\,\us_{n-3}^\top,\us_{n-2}^\top,\us_{n-1}^\top,\us_{n}^\top\,]^\top\in\R^{12}\eqand\\
			\ux^\ttr_k&=[\,\ude_{n-3}^\top,\ude_{n-2}^\top,\ude_{n-1}^\top,\ude_{n}^\top\,]^\top\in\R^{12}\\
		\end{aligned}
	\end{equation}
	comprising the position recurrent control points (RCPs) in \eqref{eq:st} and increments of orientation RCPs in \eqref{eq:ude}. $k$ denotes the time step in state-space modeling, at which overall $n$ knots are present to span the whole spline trajectory. The continuous-time 6-DoF motion trajectory $\spx(t)$ in \eqref{eq:spl} is then established for $t\in[\,t_{n-1},t_{n})$ according to \eqref{eq:st} and \eqref{eq:rt}, which is further embedded to the state-space model as follows
	\begin{equation}\label{eq:tris}
		\begin{aligned}
			\ux_{k+1}&=\fA_k\ux_k+\uw_k\\
			\uz_{k}&=h\big(\spx(\ux_k;t_k)\big)+\uv_k\,.
		\end{aligned}
	\end{equation}
	The state vector $\ux_k\in\R^{24}$ is defined in \eqref{eq:rcp}. We propose a linear process model for system propagation, where the transition matrix $\fA_k\in\R^{24\times24}$ is kept to be constant and can be configured according to the specific use case. $\uz_k$ denotes the sensor measurement. The nonlinear observation function $h(\spx(\ux_k;t_k))$ maps the discrete-time state to the measurement domain through kinematic interpolation at timestamp $t_k$ according to \secref{subsec:kinoItp}. See \figref{fig:spline} for the conceptual illustration. Furthermore, $\uw_k$ and $\uv_k$ denote additive process and measurement zero-mean noise terms with respective covariances $\fQ$ and $\fR$. Note that we use the orientational RCP increments $\{\ude_i\}_{i=n-3}^n$ (rather than the RCPs themselves) in the state vector \eqref{eq:rcp}. Compared with the common error-state formulation, this strategy mitigates the overall nonlinearity in estimating orientations using B-splines, thereby enabling efficient state estimation through the iterated EKF developed in \secref{subsec:rbe}. Additionally, the increments can be directly applied for kinematic interpolations in \eqref{eq:rt} and the corresponding Jacobian computations. As a result, both the methodological conciseness and computational efficiency of the proposed  B-spline-based state estimator are enhanced.
	
	\subsection{Jacobians w.r.t.\ the State Vector}\label{subsec:jac}
	Given the 6-DoF TriS model proposed in \secref{subsec:6tris}, we further provide the Jacobians of the B-spline kinematics $\rg{\spx}(t)$ w.r.t.\, the state components in \eqref{eq:rcp} to facilitate recursive estimation. According to \eqref{eq:stKino}, the Jacobian of the position kinematics $\rg{\sps}(t)$ is given by $\lpad{\rg{\sps}(t)}{\ux^\tts_k}=\rg{\blam}$ for temporal derivatives up to the second order. The Jacobian of orientation spline $\spr(t)$ w.r.t.\ orientation state $\ux_k^\ttr$ in \eqref{eq:rcp} follows
	\begin{equation}\label{eq:jacr}
		\pad{\spr(t)}{\ux^\ttr_k}=\Big[\pad{\spr(t)}{\ude_{n-3}},\pad{\spr(t)}{\ude_{n-2}},\pad{\spr(t)}{\ude_{n-1}},\pad{\spr(t)}{\ude_{n}}\Big]\in\R^{4\times12}\,,
	\end{equation}
	with each block matrix being the Jacobian w.r.t.\ the increment of orientation RCPs given by
	\begin{equation*}
		\pad{\spr{(t)}}{\ude_i}=\lambda_i\,\llmat{\fQ}_\la\,\lrmat{\fQ}_\ra\,\pad{\Exp_\dso(\unu)}{\unu}\bigg\vert_{\unu=\lambda_i\ude_i}\,,\quad i=n-3,\cdots,n\,.
	\end{equation*}
	For brevity, we exploit the substitutions  
	\begin{equation*}
		\llmat{\fQ}_\la=\textstyle\llmat{\qf}\big(\ur_{n-4}\otimes\prod_{j=n-3}^{i-1}\ue_j\big)\,\,\text{and}\,\,\lrmat{\fQ}_\ra=\lrmat{\qf}\big(\prod_{j=i+1}^{n}\ue_j\big)\,,
	\end{equation*}
	where $\llmat{\qf}(\cdot)$ and $\lrmat{\qf}(\cdot)$ denote the left and right matrix expressions of quaternion. Derivation of the partial derivative $\lpad{\Exp_\dso(\unu)}{\unu}$ is provided in~\cite[Eq.~(19)]{RAL23_Li}. Furthermore, the Jacobian of the angular velocity function $\uom(t)$ in \eqref{eq:omrec} follows
	\begin{equation}\label{eq:jacom}
		\pad{\uom(t)}{\ux^\ttr_k}=\bigg[\pad{\uom(t)}{\ude_{n-3}},\pad{\uom(t)}{\ude_{n-2}},\pad{\uom(t)}{\ude_{n-1}},\pad{\uom(t)}{\ude_{n}}\bigg]\in\R^{3\times12}\,,
	\end{equation}
	where the terms are derived as
	\begin{equation*}
		\begin{aligned}
			\pad{\uom(t)}{\ude_{n-2}}&=\pad{\uom(t)}{\uom_1(t)}\pad{\uom_1(t)}{\ude_{n-2}}=2\dot{\lambda}_{n-2}\scR(\ue_n^{-1})\scR(\ue_{n-1}^{-1})\,,\\
			\pad{\uom(t)}{\ude_{n-3}}&=\fzero_3\,,\pad{\uom(t)}{\ude_n}=\mathscr{J}_n(\uom_2(t))\,,\\
			\pad{\uom(t)}{\ude_{n-1}}&=\pad{\uom(t)}{\uom_2(t)}\pad{\uom_2(t)}{\ude_{n-1}}=\scR(\ue_n^{-1})\mathscr{J}_{n-1}(\uom_1(t))\,,\eqwith\\
			\mathscr{J}_i(\uv)&=\lambda_i\pad{(\ue_i^{-1}\bullet\uv\bullet\ue_i)}{\ue_i}\pad{\ue_i}{(\lambda_i\ude_i)}+2\dot{\lambda}_i\fI_3\,,\ i=n-1,n\,.
		\end{aligned}
	\end{equation*}
	The function $\scR(\cdot)$ maps a quaternion to its corresponding rotation matrix. The partial derivative $\lpad{(\ue_i^{-1}\bullet\uv\bullet\ue_i)}{\ue_i}$ can be computed according to~\cite[Eq.~(174)]{sola2017quaternion}.
	\begin{algorithm}[t]
		\caption{Recursive Spline Estimator (RESPLE)}\label{alg:resple}
		\KwIn{previous posterior $\hux_{k-1|k-1}$, $\fP_{k-1|k-1}$, measurement $\uz_k$ at timestamp $t_k^\ttt{z}$, maximum iteration $n_\text{max}$, convergence threshold $\epsilon$}
		\KwOut{posterior estimate $\hux_{k|k}$, $\fP_{k|k}$}
		\tcc{\color{blue}Prediction\color{black}}
		\If{$t_k^\ttt{z}<t_{n}$}{
			$\fA_{k-1}\leftarrow\fI_{24}$\,\tcp*{\color{blue}random walk}
		} \Else {
			$\fA_{k-1}\leftarrow\fA$\,\tcp*{\color{blue}knot extension}
		}
		$\hux_{k|k-1}\leftarrow\fA_{k-1}\hux_{k-1|k-1}$\,;\\
		$\fP_{k|k-1}\leftarrow\fA_{k-1}\fP_{k-1|k-1}\fA_{k-1}^\top+\fQ_{k-1}$\,;\\
		\tcc{\color{blue}Iterated Update\color{black}}
		$j\leftarrow0$\,,
		$\hux_j\leftarrow\hux_{k|k-1}$\,;\\
		\While{\text{true}}{
			$\underline{\gamma}_j\leftarrow\uz_k-h(\hux_j;t_k^\ttt{z})$\,;\\
			$\fH_j\leftarrow\texttt{computeJacobian}\big(\hux_j,t_k^\ttt{z}\big)$\,;\\
			\If{$\dim(\uz_k)\leq\dim(\hux_{k|k-1})$}{
				$\fK_j\leftarrow\fP_{k|k-1}\fH_j^\top\big(\fH_j\fP_{k|k-1}\fH_j^\top+\fR_k\big)^{-1}$\,;\\
			} \Else {
				$\fK_j\leftarrow\big(\fH_j^\top\fR_k^{-1}\fH_j+\fP_{k|k-1}^{-1}\big)^{-1}\fH_j^\top\fR_k^{-1}$\,;\\
			}
			$\delta\ux_j\leftarrow\fK_j\underline{\gamma}_j-(\fI-\fK_j\fH_j)(\hux_j-\hux_{k|k-1})$\,;\\
			$\hux_{j+1}\leftarrow\hux_j+\delta\ux_j$\,;\\
			\If{$\Vert\delta\ux_j\Vert<\epsilon$ \bf{or} $j+1=n_\text{max}$}{
				\text{break\,;}
			}
			$j\leftarrow{j}+1$\,;\\
		}
		$\hux_{k|k}\leftarrow\hux_{j+1}$\,;\\
		$\fP_{k|k}\leftarrow(\fI-\fK_j\fH_j)\fP_{k|k-1}$\,;	\\
		\Return $\hux_{k|k}\,,\fP_{k|k}$
	\end{algorithm}
	
	\subsection{Recursive Bayesian Estimation on 6-DoF B-Splines}\label{subsec:rbe}
	Based on the 6-DoF TriS model proposed in \eqref{eq:tris}, we now establish the recursive spline estimator by modifying the iterated EKF, as outlined in \algref{alg:resple} and elaborated below. 
	\subsubsection{Prediction}
	Upon receiving a new measurement $\uz_k$, we compute the predicted prior mean and covariance as
	\begin{equation}\label{eq:pred}
		\begin{aligned}
			\hux_{k|k-1}&=\fA_{k-1}\hux_{k-1|k-1}\,,\\
			\fP_{k|k-1}&=\fA_{k-1}\fP_{k-1|k-1}\fA_{k-1}^\top+\fQ_{k-1}\,,
		\end{aligned}
	\end{equation}
	with $\hux_{k-1|k-1}$ and covariance $\fP_{k-1|k-1}$ being the previous posterior mean and covariance, respectively.
	The transition matrix $\fA_{k-1}$ is selected according to the measurement timestamp. If $\uz_k$ falls within the current spline time span $t_{n}$, the knots remain the same number by setting $\fA_{k-1}=\fI_{24}$ as a random walk. Otherwise, we add a new control point to extend the current time span to $t_n+\tau$ by using a non-identity transition matrix $\fA_{k-1}=\fA$, which will be specified in \secref{subsec:rp} for LiDAR-based odometry.
	
	\subsubsection{Iterated Update}
	The iterations within the update step are initialized using the predicted prior, i.e., $\hux_j=\hux_{k|k-1}$ for $j=0$. 
	At each iteration, we compute the observation function's Jacobian $\fH_j$ at $t_k^\ttt{z}$ w.r.t.\ current RCPs via the chain rule composing sensor-specific model and the Jacobians of spline kinematics given in \secref{subsec:jac}. The current state estimate can be updated according to $\hux_{j+1}=\hux_j+\delta\ux_j$, with the increment $\delta\ux_j$ given by
	\begin{equation*}
		\begin{aligned}
			&\delta\ux_j=\fK_j(\uz_k-h(\hux_j))-(\fI-\fK_j\fH_j)(\hux_j-\hux_{k|k-1})\,,\\	
			&\text{with}\quad\fK_j=\fP_{k|k-1}\fH_j^\top\big(\fH_j\fP_{k|k-1}\fH_j^\top+\fR_k\big)^{-1}
		\end{aligned}
	\end{equation*}
	denoting the standard Kalman gain at the $j$-th iteration, and $\fR_k$ the covariance matrix of measurement noise. In the case of high-dimensional measurement $\uz_k$ (higher than the state vector), the Kalman gain from~\cite{xu2022fast} is adopted, namely,
	\begin{equation*}
		\fK_j=\big(\fH_j^\top\fR_k^{-1}\fH_j+\fP_{k|k-1}^{-1}\big)^{-1}\fH_j^\top\fR_k^{-1}\,.
	\end{equation*}
	This avoids the inversion of the high-dimensional matrix associated with the measurement space, while leveraging the block-diagonal structure of $\fR_k^{-1}$ for efficient computation. The iteration terminates when the increment $\Vert\delta\ux_j\Vert$ is sufficiently small or the maximum iteration is reached, yielding the posterior mean and covariance
	$\hux_{k|k}=\hux_{j+1}$ and $\fP_{k|k}=(\fI-\fK_j\fH_j)\fP_{k|k-1}$, respectively. 
	
	\section{LiDAR-based Odometry using RESPLE}\label{sec:rblo}
	We now customize the proposed RESPLE framework to egomotion estimation in a generic multi-LiDAR-inertial setting. The following state vector is set up accordingly 
	\begin{equation}\label{eq:liostate}
		\ux_k=\big[\,(\ux_k^\tts)^\top\,,(\ux_k^\ttr)^\top\,,\ub_\acce^\top\,,\ub_\gyro^\top\,\big]^\top\in\R^{30}\,,
	\end{equation}
	where $\ux_k^\tts$ and $\ux_k^\ttr$ are the RCP components defined in \eqref{eq:rcp} for representing the IMU body spline trajectory w.r.t.\ world frame. $\ub_\acce$ and $\ub_\gyro$ denote accelerometer and gyroscope biases within the time span of RCPs, respectively.
	\subsection{System Pipeline}
\begin{figure}[t]
		\vspace{1mm}
		\centering
		{\includegraphics[width=1.0\linewidth]{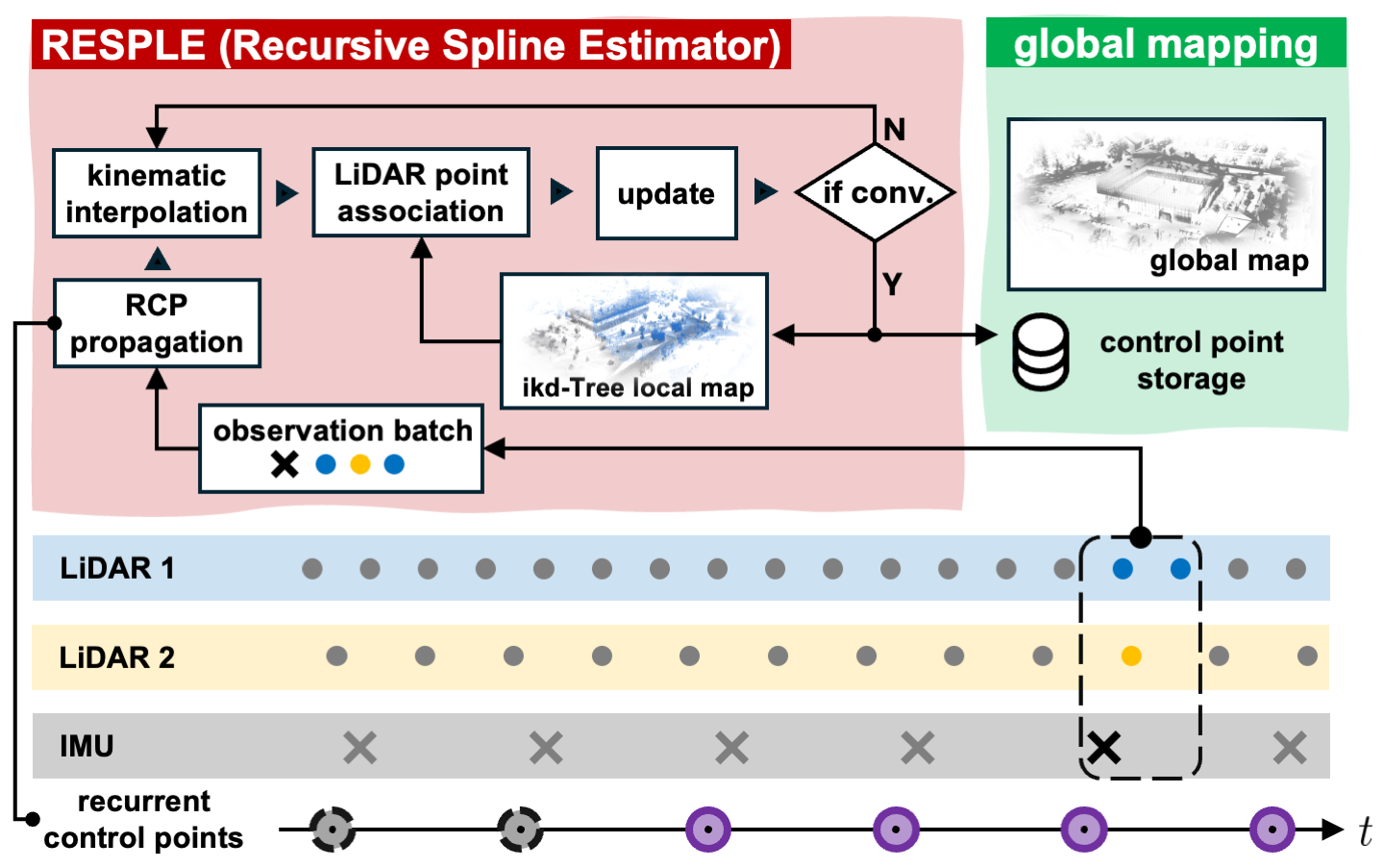}}
		\caption{RESPLE-based multi-LiDAR-inertial odometry.}
		\label{fig:resple}
		\vspace{-2mm}
	\end{figure}

	The proposed RESPLE-based multi-LiDAR-inertial odometry system is illustrated in \figref{fig:resple}. Given the multi-LiDAR input, point clouds are first downsampled using voxel grids and, together with IMU readings, queued into an observation batch $\{\uz_i^\circ\}_{i=1}^m$ according to their exact timestamps $\{t_i\}_{i=1}^m$. The superscript $\circ$ here is an umbrella term for LiDAR (\ttt{L}) and IMU (\ttt{I}) observations. The observation batch size is bounded by a predefined threshold and the time span of the latest knot. Depending on the latest measurement's timestamp, we perform prediction on RCPs through either extension or random walk (\secref{subsec:rbe}). We further perform kinematic interpolations at the exact timestamps of multi-sensor data points. Each LiDAR point is retrieved within the world frame without explicit de-skewing, followed by  association to a spatial local map managed by the ikd-Tree~\cite{xu2022fast}. Once the measurement model in \eqref{eq:tris} is established, we perform iterated update to obtain posterior estimates of the RCPs. As time progresses, active RCPs transition into idle state, and corresponding LiDAR points are interpolated for maintaining the local map as well as the global trajectory and map.
	
	\subsection{RESPLE Prediction}\label{subsec:rp}
	We now concretize the non-identity transition matrix $\fA$ for \eqref{eq:pred} in the case of knot extension. Various kinematic principles can be applied to RCP propagation; here, we adopt a straightforward strategy similar to~\cite{li2025embedding} that preserves the translational and angular velocities of a preceding RCP at the newly added knot. This yields the block-diagonal transition matrix $\fA=\diag(\fA_\tts,\fA_\ttr,\fI_3,\fI_3)$, with the translational and rotational submatrices as follows 
	\begin{equation*}
		\fA_\tts=\Bigg[\sbbmat \fzero_3 &\fI_3&\fzero_3&\fzero_3&\\
		\fzero_3 &\fzero_3 &\fI_3 &\fzero_3\\\fzero_3&\fzero_3&\fzero_3&\fI_3\\
		-\fI_3 &\fzero_3 &2\fI_3 &\fzero_3\sebmat\Bigg]\eqand\fA_\ttr=\Bigg[\sbbmat \fzero_3 &\fI_3&\fzero_3&\fzero_3&\\
		\fzero_3 &\fzero_3 &\fI_3 &\fzero_3\\\fzero_3&\fzero_3&\fzero_3&\fI_3\\
		\fzero_3 &\fI_3 &\fzero_3 &\fzero_3\sebmat\Bigg]\,.
	\end{equation*}	
	
	\subsection{RESPLE Update}\label{subsec:ru}
	The basic design in RESPLE processes sensor measurements in a point-wise manner. To ensure real-world runtime efficiency and robustness, a sequence of multi-sensor measurements are temporally stacked into an observation batch $\uz_k=[(\uz_1^\circ)^\top,\cdots,(\uz_m^\circ)^\top]^\top$ w.r.t.\ their exact timestamps $\{t_i\}_{i=1}^m$. Accordingly, we concatenate their Jacobians temporally for iterated update, i.e., $\fH_j=[(\fH_1^\circ)^\top,\cdots,(\fH_m^\circ)^\top]^\top$, with $j$ denoting the iteration index. The LiDAR and IMU measurement models are concretized as follows.
	
	\subsubsection{LiDAR point-to-plane metric}
	Given a LiDAR point observed at timestamp $t_i$ and its coordinates $\up_i^\flid$ in the LiDAR frame, we establish an observation function for \eqref{eq:tris} based on point-to-plane distance as follows
	\begin{equation}\label{eq:zL}
		{h}_i^\ttt{L}(\ux_k,t_i)=\un_i^\top\big(\scR(t_i)\,\up_i^\fimu+\sps(t_i)-\ual_i\big)\,.
	\end{equation}
	$\up_i^\fimu=\fR_\flid^\fimu\,\up_i^\flid+\us_\flid^\fimu$ transforms the point from LiDAR to IMU body frame through the extrinsic $\fR_\flid^\fimu\in\SOT$ and $\us_\flid^\fimu\in\R^3$. This point is further transformed to world frame through interpolation on the 6-DoF spline trajectory in \eqref{eq:spl} at $t_i$, where the rotation matrix $\scR(t_i)\in\SOT$ is applied given quaternion $\spr(t_i)$. We further associate it to the nearest neighbor $\ual_i$ in the local map through ikd-Tree~\cite{xu2022fast}, where a plane is fitted using $N$ points (e.g., $N=5$) in the vicinity. The point-to-plane distance is then computed with the normal vector $\un_i$ of the associated plane. Correspondingly, the Jacobian of \eqref{eq:zL} w.r.t.\ the state vector \eqref{eq:liostate} is derived as  
	\begin{equation*}
		\fH_i^\ttt{L}=\bbmat\un_i^\top\blam\,,\,\un_i^\top\pad{\scR(t_i)\,\up_i^\fimu}{\spr(t_i)}\pad{\spr(t_i)}{\ux^\ttr_k}\,,\uzero_3^\top\,,\uzero_3^\top\ebmat\in\R^{1\times30}\,,
	\end{equation*}
	where $\lpad{\spr(t_i)}{\ux^\ttr_k}$ is provided in \eqref{eq:jacr}. For outlier rejection, we require the metric's variance estimate $\fH_i^\ttt{L}\fP_{k\vert{k}-1}(\fH_i^\ttt{L})^\top+\fR^\ttt{L}$ to be below a predefined threshold. Here, $\fR^\ttt{L}$ denotes the LiDAR noise variance, and $\fP_{k\vert{k}-1}$ is the prior covariance obtained directly from the RESPLE prediction. 
	
	\subsubsection{IMU metric}
	Suppose an IMU measurement $\uz_i^\ttt{I}=[(\uz_i^\acce)^\top,(\uz_i^\gyro)^\top]^\top$ is received at timestamp $t_i$ in batch $\uz_k$, comprising both accelerometer and gyroscope readings. The observation model in \eqref{eq:tris}  is then specified as
	\begin{equation}\label{eq:mmi}
		{h}_i^\ttt{I}(\ux_k,t_i)=\bbmat\scR(t_i)^\top(\ddot{\sps}(t_i)+\ug)+\ub_{\acce}\\\uom(t_i)+\ub_{\gyro}\ebmat\,,
	\end{equation}
	where $\ddot{\sps}(t_i)$ and $\uom(t_i)$ denote the acceleration and angular velocity at $t_i$, expressed in the world and body frames, respectively, according to kinematic interpolations \eqref{eq:stKino} and \eqref{eq:omrec}. The acceleration $\ddot{\sps}(t_i)$ is then combined with the gravity vector $\ug$ and transformed to the IMU body by $\scR(t_i)\in\SOT$ obtained via \eqref{eq:rt}. Furthermore, we provide the Jacobian of \eqref{eq:mmi} w.r.t.\ the state vector \eqref{eq:liostate} as follows
	\begin{equation*}
		\setlength{\arraycolsep}{2pt}
		\fH_i^\ttt{I}=\bbmat\scR(t_i)^\top\ddot{\blam}&\pad{\scR(t_i)^\top(\ddot{\sps}(t_i)+g)}{\spr(t_i)}\pad{\spr(t_i)}{\ux^\ttr_k}&\fI_3&\fzero_3\\
		\fzero_{3\times12}&\pad{\uom(t_i)}{\ux^\ttr_k}&\fzero_3&\fI_3\ebmat\in\R^{6\times{30}}\,,
	\end{equation*}
	where $\lpad{\uom(t_i)}{\ux^\ttr_k}$ is given in \eqref{eq:jacom}. 
	\subsection{Implementation}
	The proposed LiDAR-based odometry system is developed in C++ as a ROS2 package, with Eigen for linear algebra operations~\cite{macenski2022robot,eigenweb}. As illustrated in \figref{fig:resple}, our system comprises two ROS nodes: the recursive spline estimator (RESPLE), including data preprocessing and association, and the global mapping module. Within the RESPLE node, we exploit OpenMP~\cite{dagum1998openmp} for parallelizing kinematic interpolations (including calculating Jacobians) and LiDAR point associations. Our software package is designed to support a variety of LiDAR-based multi-sensor settings for LiDAR (LO), LiDAR-inertial (LIO), multi-LiDAR (MLO) and multi-LiDAR-inertial (MLIO) odometry, sharing RESPLE as the core algorithm for motion estimation. 
	
	\section{Evaluation}\label{sec:eva}
	We conduct extensive real-world benchmarking involving public datasets and own experiments. All evaluations are conducted on a laptop running Ubuntu 22.04 (Intel i7-11800H CPU, 48GB RAM).
	
	\subsection{Benchmarking Setup}
	We include public datasets \ttt{NTU VIRAL}~\cite{viralData}, \ttt{MCD}~\cite{nguyen2024mcd}, and \ttt{GrandTour}~\cite{frey2025boxi}, and our own experimental dataset \ttt{HelmDyn} for evaluations, as described in Tab.~\ref{tab:dataset} for various mobile platforms, scenarios, and sensor configurations. RESPLE (R) is  consistently compared against state-of-the-art systems: Traj-LO (T-LO)~\cite{zheng2024trajlo}, CTE-MLO (C-MLO)~\cite{shen2025cte}, and FAST-LIO2 (F-LIO2)~\cite{xu2022fast}. Abbreviations in parentheses are used for brevity. We configure RESPLE with a knot frequency of \SI{100}{\hertz} and a maximum of 5 iterations in iterated update. In each dataset, the same parameter set is used without individual tuning, where the observation batch spans \SI{3}{} to \SI{10}{\ms}. For accuracy quantification, we interpolate trajectory estimates  at timestamps of ground truth and compute RMSE of the absolute position error (APE) using \textit{evo}~\cite{grupp2017evo} except for \ttt{NTU VIRAL} (official evaluation script is used instead).  We mark failures using \xmark, and the best and second-best results with \textbf{bold} and \underline{underline}, respectively.
	\begin{table}[h!]
		\vspace{-1mm}
		\centering
		\caption{Datasets for real-world benchmarking.}
		\begin{tabular}{lll}
			\toprule 
			\textbf{Dataset} & \textbf{Scenarios} & \textbf{LI Sensors (Adopted)}\\
			\midrule
			\rowcolor{gray!15} 
			\ttt{NTU VIRAL}~\cite{viralData}  &indoor, outdoor, &{Ouster OS1-16} (\textbf{L})\\
			\rowcolor{gray!15} 
			&drone &{VN100} \textbf{(I)} \\
			\ttt{MCD}~\cite{nguyen2024mcd}  & large-scale urban, &{Livox Mid70} (\textbf{L}) \\
			&fast, ground vehicle &{VN100} (\textbf{I}) \\
			\rowcolor{gray!15} 
			\ttt{GrandTour}~\cite{frey2025boxi} & wild, urban,  & {Hesai XT32} (\textbf{L1}) \\
			\rowcolor{gray!15} &underground &{Livox Mid360} (\textbf{L2}) \\
			\rowcolor{gray!15} &quadruped robot &{built-in L2} (\textbf{I})\\
			\ttt{HelmDyn} &indoor, dynamic, &{Livox Mid360} (\textbf{L}) \\
			(own experiment) &wearable (helmet)  &{built-in L} (\textbf{I})  \\
			\bottomrule
		\end{tabular}
		\label{tab:dataset}
	\end{table}
	
	\subsection{Public Datasets}
	\subsubsection{\ttt{NTU VIRAL}}
	We adopt the horizontal LiDAR~\cite{viralData} for RESPLE. Shown in Tab.~\ref{tab:ntumcd}, our LO/LIO systems consistently rank among the top two in accuracy across all sequences without any failures, and overall outperforms CTE-MLO (using 2 LiDARs) and FAST-LIO2.
	
	\subsubsection{\ttt{MCD}}
	We select 6 fast, large-scale sequences, covering both day and night scenarios~\cite{nguyen2024mcd} listed in Tab.~\ref{tab:ntumcd}. RESPLE-based LO/LIO systems consistently deliver comparable estimation accuracy to state-of-the-art methods, without encountering any failures.
	
	\begin{table}[htbp]
		\vspace{1mm}
		\centering
		\setlength{\tabcolsep}{4pt}
		\caption{APE (RMSE, meters) on \ttt{NTU VIRAL} and \ttt{MCD}.}
			\begin{tabular}{@{}r|ccc|cc@{}}
				\toprule
				\ttt{\textbf{NTU VIRAL}} &T-LO${}^1$ &C-MLO${}^2$ &F-LIO2${}^3$ &{R-LO} &{R-LIO}\\
				\midrule
				\ttt{eee\_01} &${0.055}$ &$0.08$ &$0.069$ &$\underline{0.044}$ &$\bf{0.036}$\\
				\ttt{eee\_02} &$0.039$ &$0.07$ &$0.069$ &$\underline{0.023}$ &$\bf{0.022}$\\
				\ttt{eee\_03} &$\underline{0.035}$ &$0.12$ &$0.111$ &$0.046$ &$\bf{0.033}$\\
				\ttt{nya\_01} &$0.047$ &$0.06$ &$0.053$ &$\underline{0.033}$ &$\bf{0.030}$\\
				\ttt{nya\_02} &$0.052$ &$0.09$ &$0.090$ &$\underline{0.036}$ &$\bf{0.032}$\\
				\ttt{nya\_03} &$0.050$ &$0.10$ &$0.108$ &$\underline{0.037}$ &$\bf{0.030}$\\
				\ttt{rtp\_01} &$\bf{0.050}$ &$0.13$ &$0.125$ &$0.059$ &$\underline{0.052}$\\
				\ttt{rtp\_02} &$\underline{0.058}$ &$0.14$ &$0.131$ &$0.071$ &$\bf{0.049}$\\
				\ttt{rtp\_03} &${0.057}$ &$0.14$ &$0.137$ &$\bf{0.054}$ &$\underline{0.056}$\\	
				\ttt{sbs\_01}  &$0.048$ &$0.09$ &$0.086$ &$\underline{0.040}$ &$\bf{0.034}$\\
				\ttt{sbs\_02} &${0.039}$ &$0.08$ &$0.078$ &$\underline{0.034}$ &$\bf{0.031}$\\
				\ttt{sbs\_03} &$0.039$ &$0.09$ &$0.076$ &$\underline{0.036}$ &$\bf{0.033}$\\	
				\ttt{spms\_01}  &$\underline{0.121}$ &$0.21$ &$0.210$ &$\bf{0.108}$ &$0.125$\\
				\ttt{spms\_02} &\xmark &$0.33$ &$0.336$ &$\underline{0.130}$ &$\bf{0.121}$\\
				\ttt{spms\_03} &$\bf{0.103}$ &${0.20}$ &${0.217}$ &$0.216$ &$\underline{0.109}$\\
				\ttt{tnp\_01}  &$0.505$ &$0.09$ &$0.090$ &$\underline{0.052}$ &$\bf{0.049}$\\
				\ttt{tnp\_02}  &$0.607$ &$0.09$ &$0.110$ &$\underline{0.070}$ &$\bf{0.047}$\\
				\ttt{tnp\_03}  &$0.101$ &$0.10$ &$0.089$ &$\underline{0.050}$ &$\bf{0.046}$\\		
				\bottomrule
				\toprule
				\ttt{\textbf{MCD}} &T-LO & C-MLO &F-LIO2${}^3$ &{R-LO} &{R-LIO}\\
				\midrule
				\ttt{ntu\_day\_01}   &\xmark &$\underline{0.715}$ &${0.901}$ &${0.910}$ &$\bf{0.549}$\\
				\ttt{ntu\_day\_02}   &$0.194$ &$\bf{0.164}$ &$0.185$ &$\underline{0.178}$ &$0.188$\\
				\ttt{ntu\_day\_10}   &$\underline{1.129}$ &$\bf{1.112}$ &$1.975$ &${1.402}$ &$1.493$\\
				\ttt{ntu\_night\_04}   &$\underline{0.427}$ &$0.774$ &$0.902$ &${0.486}$ &$\bf{0.416}$\\
				\ttt{ntu\_night\_08}   &${0.950}$ &$\bf{0.738}$ &$1.002$ &$0.953$ &$\underline{0.940}$\\
				\ttt{ntu\_night\_13}   &\xmark &$\bf{0.461}$ &$1.288$ &$\underline{0.513}$ &${0.560}$\\		
				\bottomrule
			\end{tabular}\\
		\vspace{1mm}
		$^{1,2,3}$Results taken from \cite{zheng2024trajlo}, \cite{shen2025cte} and \cite{ng2024eigen}.
		\label{tab:ntumcd}
	\end{table}

	\subsubsection{\ttt{GrandTour}}
	The \ttt{GrandTour}~\cite{frey2025boxi} is a new legged robotics dataset of immense scale and diversity. An ANYmal D quadruped robot equipped with a new open-source multi-sensor rig \textit{Boxi}~\cite{frey2025boxi} traversed 71 Swiss environments under diverse conditions, covering a total of \SI{15}{\kilo\meter} over 8 hours. As listed in Tab.~\ref{tab:grandtour}, we select 2 sequences recorded underground (\ttt{JTL/S}), 1 urban sequence (\ttt{HEAP-1}) and 5 in the wild (forests and mountains). The sequences present challenges due to dynamic motions and cluttered or geometrically degenerate scenes. Traj-LO, CTE-MLO and FAST-LIO2 exhibit multiple failures. Our LiDAR-only variant performs well with only one failure. Moreover, adding an additional LiDAR and IMU within RESPLE can significantly improve estimation accuracy and robustness. An exemplary run of RESPLE-MLO on \ttt{ALB-2} is illustrated in \figref{fig:front}-(B).
	\begin{table}[htbp]
		\setlength{\tabcolsep}{3.2pt} 
		\caption{APE (RMSE, meters) on \ttt{GrandTour}.}
		\begin{tabular}{@{}r|ccc|cccc@{}}
			\toprule
			&T-LO &C-MLO &F-LIO2 &{R-LO} &{R-MLO} &{R-LIO} &{R-MLIO}\\
			&\footnotesize{L1}   &L1+L2  &L1+I   &L1     &L1+L2   &L1+I    &\footnotesize{L1+L2+I}\\
			\midrule
			\ttt{JTL}$^*$   &$0.046$ & \xmark &\xmark &${0.035}$ &$\bf{0.026}$ &$\underline{0.028}$ &$\underline{0.028}$ \\
			\ttt{JTS}$^*$   &$\bf{0.074}$ & $0.264$ &$2.585$ &$0.256$ &$\underline{0.088}$ &$0.128$ &${0.091}$ \\
			\ttt{RIV-1}  &\xmark & \xmark &$5.522$ &$\underline{0.041}$ &${0.066}$ &$\bf{0.039}$ &$0.046$ \\
			\ttt{PKH}$^*$  &\xmark & $4.576$ &\xmark &\xmark &$\underline{0.059}$ &${0.076}$ &$\bf{0.048}$ \\
			\ttt{HEAP-1}  &$0.045$ & $0.038$ &$0.151$ &$\underline{0.026}$ &${0.027}$ &$\bf{0.022}$ &${0.028}$ \\		
			\ttt{TRIM-1}  &$0.047$ & $0.063$ &$0.218$ &${0.039}$ &$\underline{0.030}$ &$0.037$ &$\bf{0.028}$ \\
			\ttt{ALB-2}  &$0.044$ & $0.050$ &$0.442$ &$\underline{0.018}$ &$0.029$ &$\bf{0.015}$ &$0.031$ \\		
			\ttt{LMB-2}  &$0.043$ & \xmark &$3.086$ &$0.040$ &$\underline{0.039}$ &$0.046$ &$\bf{0.035}$ \\	
			\bottomrule
		\end{tabular}
		\vspace{1mm}
		\\
		$^*$Not available in public release.
		\label{tab:grandtour}
	\end{table}
	
	\subsection{Own Experiments}
	\subsubsection*{\ttt{HelmDyn} (Helmet Dynamic)}
\begin{figure}[t!]
		\vspace{1mm}
		\begin{tabular}{cc}
			\adjustbox{trim={0.\width} {0.\height} {0.\width} {0.\height},clip}{\includegraphics[width=0.29\textwidth]{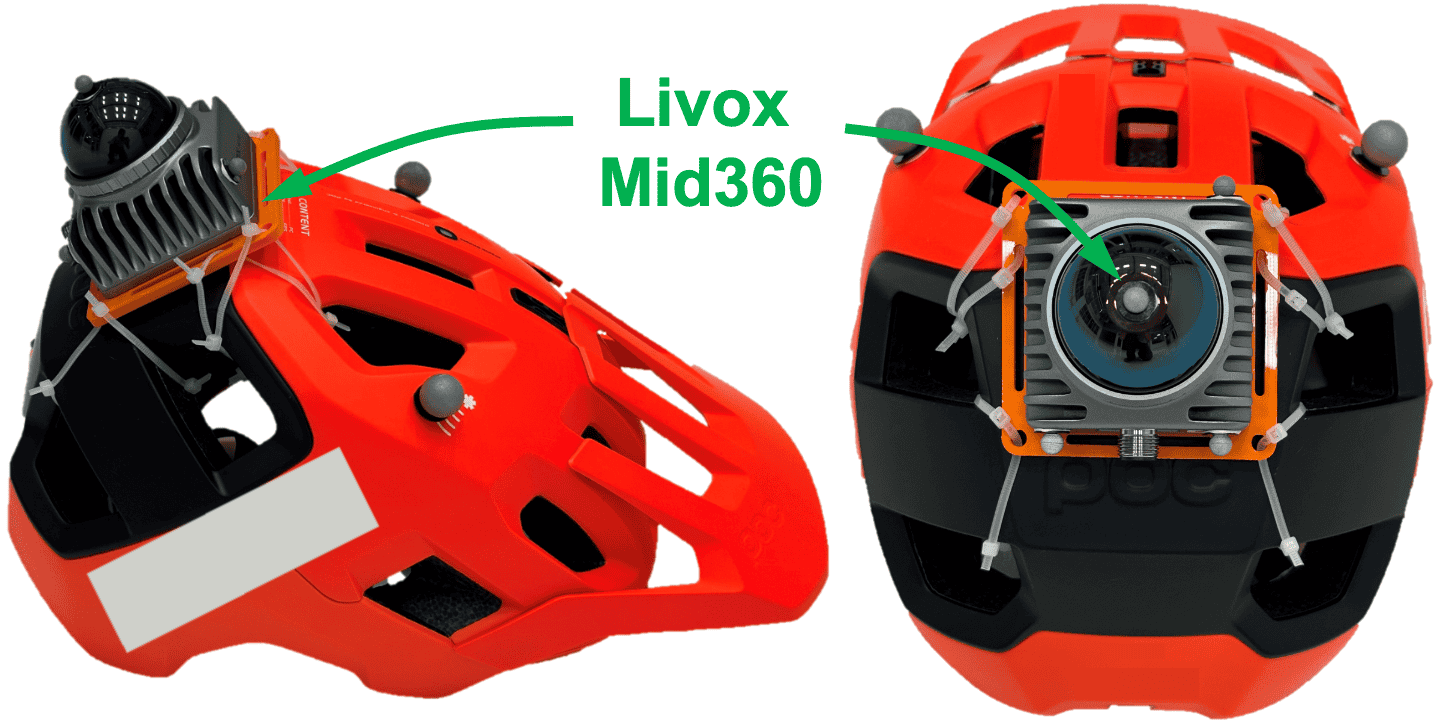}}&
			\adjustbox{trim={0.\width} {0.\height} {0.\width} {0.\height},clip}{\includegraphics[width=0.1\textwidth]{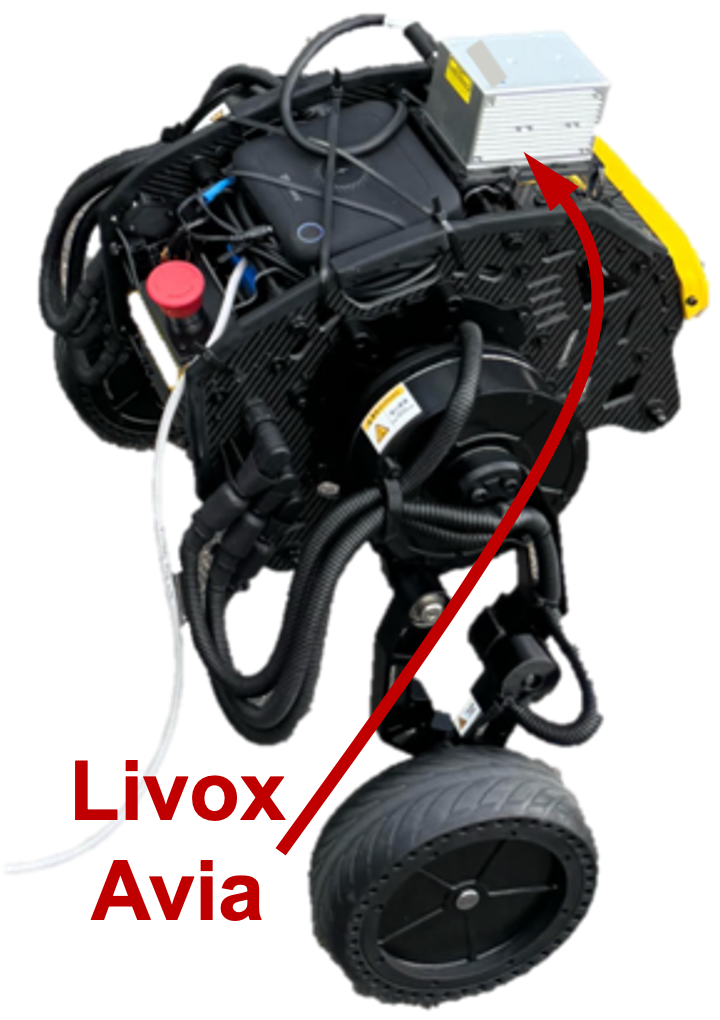}}\\
			\subcap{(A) \ttt{HelmDyn}} &\subcap{(B) \ttt{R-Campus}}\\
		\end{tabular}
		\caption{Mobile platforms in our experiments.}
		\label{fig:experiment}
		\vspace{-3mm}
	\end{figure}

\begin{figure}[t!]
		\vspace{1mm}
		\centering
		\begin{tabular}{cc}
			\includegraphics[width=0.225\textwidth]{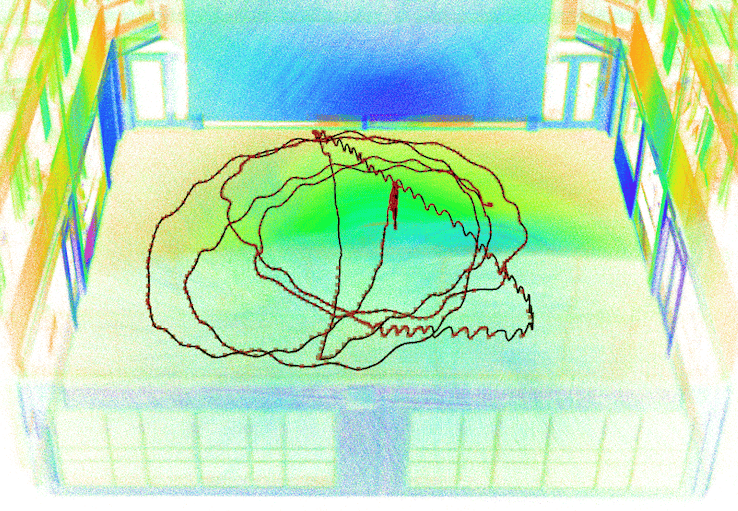}&
			\includegraphics[width=0.225\textwidth]{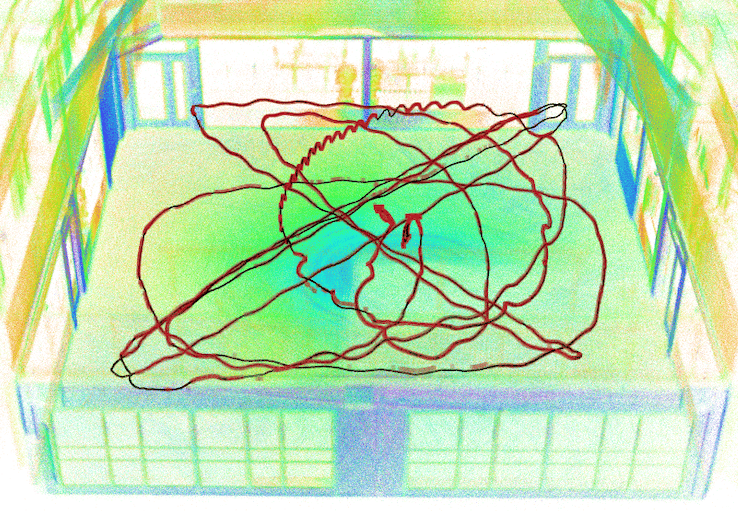}\\
			\subcap{\ttt{HD\_03}} &\subcap{\ttt{HD\_06}}\\
			\includegraphics[width=0.225\textwidth]{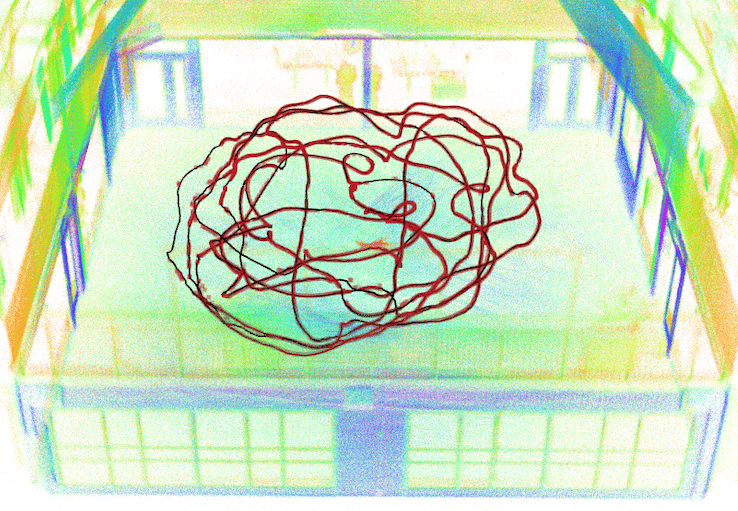}&
			\includegraphics[width=0.225\textwidth]{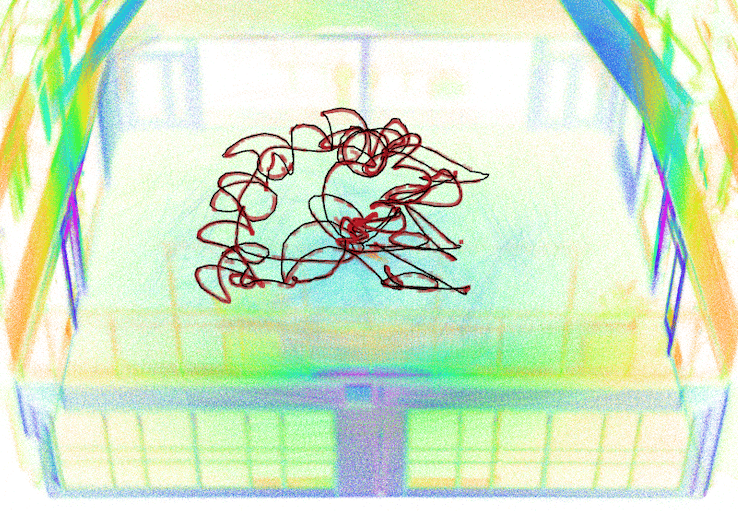}\\
			\subcap{\ttt{HD\_09}} &\subcap{\ttt{HD\_10}}
		\end{tabular}
		\caption{RESPLE-LO tested on \ttt{HelmDyn}. Black and red curves are estimate and ground truth, respectively.}
		\label{fig:helmdyn}
		\vspace{0mm}
	\end{figure}

	We conduct our own experiments using a helmet-mounted Livox Mid360, shown in \figref{fig:experiment}-(A), operated in a $12\times12\times8$\,\SI{}{\meter^3} cubic space along with dynamic movements combining walking, running, jumping, and in-hand waving. Ground truth trajectories are acquired using a high-precision (submillimeter), low-latency motion capture system consisting of 12 Oqus 700+ and 8 Arqus A12 Qualisys cameras with passive markers. As shown in Tab.~\ref{tab:helmdyn}, we additionally include Point-LIO (P-LIO) and SLICT2 (B-spline-based LIO using sliding-window optimization) due to their potential capabilities in estimating aggressive motions~\cite{he2023point,ng2024eigen}. Across the entire \ttt{HelmDyn} dataset, RESPLE consistently outperforms state-of-the-art methods despite dynamic motions.  A few exemplary runs of RESPLE-LO are illustrated in \figref{fig:front}-(C) and \figref{fig:helmdyn}.
	\begin{table}[htbp]               
		\centering
		\setlength{\tabcolsep}{4pt} 
		\caption{APE (RMSE, meters) on \ttt{HelmDyn}.}
		\begin{tabular}{@{}r|ccccc|cc@{}}
			\toprule
			& {T-LO} & C-MLO & {F-LIO2} &P-LIO & {SLICT2} & {R-LO} & {R-LIO} \\
			\midrule
			\ttt{HD\_01}  &$0.081$  &$0.055$ &$0.062$ & $0.109$  &$0.086$ &$\underline{0.041}$ & $\bf{0.039}$ \\
			\ttt{HD\_02}  &$3.667$  &\xmark &$0.073$  & $5.206$&$0.046$ &$\bf{0.033}$ & $\underline{0.037}$ \\
			\ttt{HD\_03}  &$0.043$  &$0.037$ &$0.033$  & $0.048$ &$0.046$ &$\bf{0.021}$ & $\bf{0.021}$ \\
			\ttt{HD\_04}  &$0.089$  &${0.041}$ &$0.054$& $0.100$ & $\underline{0.037}$ & $\underline{0.037}$ & $\bf{0.034}$ \\
			\ttt{HD\_05}  &$0.052$  &$0.033$ &$0.028$  & $0.055$ &$0.057$ &$\underline{0.021}$ & $\bf{0.020}$ \\
			\ttt{HD\_06}  &$0.059$  &$0.036$ &$0.039$  & $0.063$ &$0.057$ &$\bf{0.021}$ & $\underline{0.022}$ \\
			\ttt{HD\_07}  &$3.671$  &$0.048$ &$0.052$  & $0.066$ &$0.056$ &$\bf{0.031}$ & $\underline{0.032}$ \\
			\ttt{HD\_08}  &$0.054$  &$0.035$ &$0.030$  &$0.061$ &$0.048$ &$\bf{0.019}$ & $\underline{0.020}$ \\
			\ttt{HD\_09}  &$0.063$  &$0.042$ &$0.042$  & $0.063$ &$0.063$ &$\bf{0.025}$ & $\underline{0.026}$ \\
			\ttt{HD\_10}  &$1.616$  &\xmark &$0.063$  & $4.253$ &$0.054$ &$\bf{0.034}$ & $\bf{0.034}$ \\
			\bottomrule
		\end{tabular}
		\label{tab:helmdyn}
	\end{table}
	
	\subsubsection*{\ttt{R-Campus}} We record a sequence using a Livox Avia on a bipedal wheeled robot (DIABLO) shown in \figref{fig:experiment}-(B)~\cite{liu2024diablo}. It operates within a campus over a trajectory of approximately \SI{1400}{\meter} at \SI{1.2}{\meter/\second}. The route starts and ends at the same location. Our RESPLE-based LO and LIO achieve end-to-end errors of \SI{0.28}{\meter} and \SI{0.27}{\meter}, respectively -- better than CTE-MLO (\SI{0.30}{\meter}), FAST-LIO2 (\SI{2.70}{\meter}), and Traj-LO (\SI{80.31}{\meter}). An exemplary run is illustrated in \figref{fig:front}-(A).
	
	\subsection{Runtime Analysis}
	We now evaluate the runtime efficiency of RESPLE in various settings and compare it with other continuous-time systems. We configure the observation batch with a time span strictly equal to the knot interval (\SI{10}{\ms}). For all involved systems, multi-threading is set with 5 CPU threads. 
		
	As shown in Tab.~\ref{tab:runtime}, we select 3 representative sequences and present the average LiDAR point number and processing time of the RESPLE node (\figref{fig:resple}), including the iterated update. Our systems achieve an estimated theoretical speed of $2$x to $9$x the real-time requirement (\SI{10}{\ms}).
	\begin{table}[htbp]
		\caption{Runtime for RESPLE-based LiDAR odometry.}
		\centering
		\begin{tabular}{@{}rlc|c|c@{}}
			\toprule
			&\textbf{\#Pts} &\textbf{Settings} &\textbf{Iter. Update (\SI{}{\ms})} &\textbf{Total (\SI{}{\ms})}\\
			\midrule
			\ttt{HD\_03}  &$296$ &LO/LIO &$1.19$/$1.48$ &$1.40$/$1.74$ \\
			\ttt{eee\_01} &$184$ &LO/LIO &$0.68$/$0.82$ &$0.97$/$1.18$  \\
			\ttt{ALB-2} &$628$ &MLO/MLIO &$2.77$/$2.92$ & $4.15$/$4.46$\\
			\bottomrule
		\end{tabular}
		\label{tab:runtime}
	\end{table}
	
	Tab.~\ref{tab:runtimecmp} summarizes runtime comparisons on \ttt{HD\_03} using the runtime efficiency metric ($\xi$) in \cite{shen2025cte} defined as the ratio of processing time to the available time determined by the observation interval. $\xi\leq1$ indicates real-time efficiency. Though operating under constrained mobile computing conditions, RESPLE clearly exhibits the fastest performance.
	\begin{table}[htbp]
		\caption{Runtime comparisons on \ttt{HD\_03}.}
		\centering
		\begin{tabular}{@{}l|ccc|cc@{}}
			\toprule
			& {T-LO} & {C-MLO} & {SLICT2} & {R-LO} & {R-LIO} \\
			\midrule
			\textbf{Processing time (\SI{}{\ms})}  &$11.55$ &$7.85$ &$165.73$ &$1.40$ & $1.74$\\
			\textbf{Available time (\SI{}{\ms})} &$50$ &$10$ &$50$ &$10$ & $10$\\
			\midrule
			\textbf{Runtime efficiency $\xi$}       & $0.23$       & $0.79$       & $3.31$        & \bf{$0.14$}   &  $0.17$ \\
			\bottomrule
		\end{tabular}    
		\label{tab:runtimecmp}
	\end{table}
	
	\begin{figure}[t!]
		\centering
		\adjustbox{trim={0.08\width} {0.03\height} {0.05\width} {0.01\height},clip}{\includegraphics[width=0.55\textwidth]{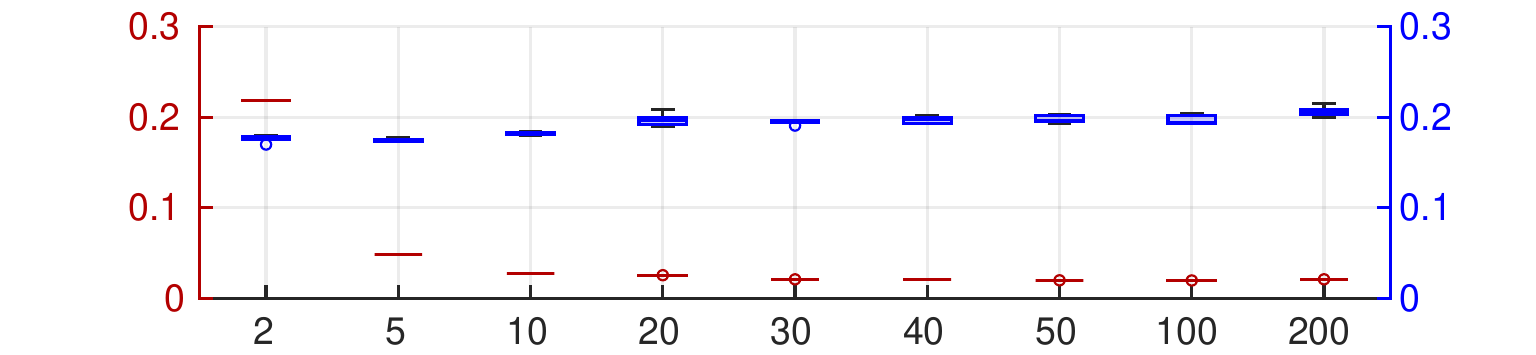}}\\
		(A) Knot frequency (batch size fixed to $100$)\\
		\adjustbox{trim={0.08\width} {0.02\height} {0.05\width} {0.01\height},clip}{\includegraphics[width=0.55\textwidth]{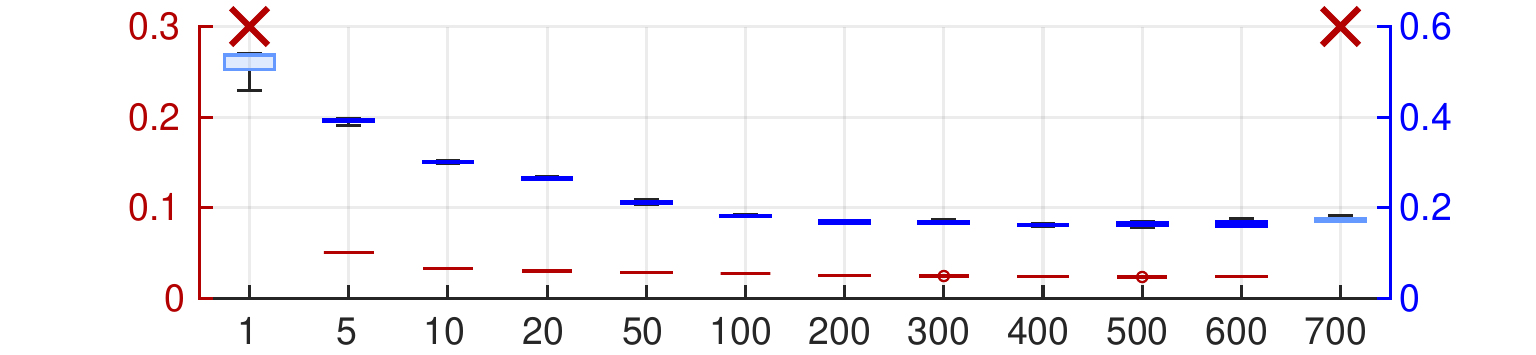}}\\
		(B) Observation batch size (knot frequency fixed to \SI{10}{\hertz})\\
		\caption{RMSEs (red, meters) and runtime efficiency $\xi$ (blue) of RESPLE-LO on \ttt{HD\_08} over varying knot frequency (A) and observation batch sizes (B). Results are plotted using \ttt{boxchart} function in \ttt{MATLAB}. {\color{kitred}$\pmb{\pmb\times}$} indicates failures.}
		\label{fig:discuss}
		\vspace{-2mm}
	\end{figure}
	
	\subsection{Parameter Analysis}
	Knot frequency and observation batch size are two key parameters in RESPLE. We investigate their impact on estimation accuracy (RMSEs) and runtime ($\xi$) in LO using \texttt{HD\_08}, with $5$ runs conducted for each configuration. As shown in \figref{fig:discuss}-(A), increasing the knot frequency yields lower estimation error, while the runtime remains approximately constant. Increasing the observation batch can improve both estimation accuracy and runtime, as illustrated in \figref{fig:discuss}-(B). However, incorporating either point-wise (batch size of $1$) or large-batch observations tends to compromise estimation robustness, primarily due to increased vulnerability to false point associations, especially under low knot frequencies.          
	
	\subsection{Discussion}    
	\subsubsection*{Estimation accuracy and robustness}
	Overall, RESPLE enables more accurate estimation than existing discrete-time systems, owing to its continuous-time motion representation. For typical aerial or wheeled platforms following relatively smooth trajectories within geometrically well-conditioned environments, RESPLE enables comparable estimation accuracy to state-of-the-art continuous-time systems. Under challenging conditions such as dynamic motions within cluttered scenes (\ttt{GrandTour} and \ttt{HelmDyn}), RESPLE outperforms existing methods in terms of accuracy and robustness, especially by adding additional LiDAR or IMU sensors. Compared to Traj-LO (constant velocity) and CTE-MLO (constant acceleration and angular velocity), the adopted cubic B-splines offer more expressive motion modeling through the piece-wise constant-jerk setting. Moreover, the proposed recursive scheme enables more frequent state propagation and update than a sliding-window spline optimization method (SLICT2), potentially improving accuracy in estimating highly dynamic motions, as shown on \ttt{HelmDyn}.
	
	\subsubsection*{Runtime efficiency and versatility} 
	The proposed recursive Bayesian scheme, including the formulation of orientational RCP increments and batch-wise update, enables a lightweight and flexible system design, delivering consistent real-time performance across diverse multi-sensor settings and scenarios. This distinguishes RESPLE from existing standalone (M)LO/LIO solutions, demonstrating strong potential as a universal motion estimator in mobile applications. 
	
	\subsubsection*{Parameter tuning} We recommend configuring RESPLE with a sufficiently high knot frequency (like \SI{100}{\hertz}) to accurately capture dynamic motion and accommodate complex environments. In parallel, selecting a reasonable large observation batch size enhances both runtime efficiency and estimation robustness, resulting in overall reliable performance for LiDAR-based odometry.

	\section{Conclusion}
	We proposed RESPLE, the first recursive 6-DoF motion estimator using B-splines. The state vector comprises position RCPs and orientation RCP increments, which are efficiently estimated through a modified iterated EKF. RESPLE further enabled a versatile, unified suite of direct LiDAR-based odometry solutions for diverse multi-sensor settings and scenarios, showing state-of-the-art or superior performance in accuracy and robustness, while attaining real-time efficiency. For future work, we will integrate visual sensors into the RESPLE pipeline to address potential degeneracy cases and incorporate a backend for global correction. Furthermore, RESPLE’s uncertainty-aware, continuous-time trajectory estimates present promising opportunities for downstream tasks such as motion planning and control.
	
	\bibliographystyle{IEEEtran.bst}
	\bibliography{bibliography.bib}
	
\end{document}